\documentclass[lettersize,journal]{IEEEtran}
\usepackage{amsmath,amsfonts}
\usepackage{algorithmic}
\usepackage{algorithm}
\usepackage{array}
\usepackage[caption=false,font=normalsize,labelfont=sf,textfont=sf]{subfig}
\usepackage{textcomp}
\usepackage{stfloats}
\usepackage{url}
\usepackage{verbatim}
\usepackage{graphicx}
\usepackage{cite}
\usepackage{multirow}
\usepackage{amsfonts}
\usepackage{amsmath}
\usepackage{arydshln}
\usepackage{url}
\usepackage{hyperref}
\usepackage{booktabs}   
\usepackage{multirow}   
\usepackage{array}
\usepackage{graphicx}
\usepackage{tcolorbox} 

\begin{document}

\title{A Hybrid Approach for Closing the Sim2real Appearance Gap in Game Engine Synthetic Datasets}

\author{Stefanos Pasios
\thanks{The author is an independent researcher (email: stefanos.pasios@gmail.com).}}



\maketitle

\begin{abstract}
Video game engines have been an important source for generating large volumes of visual synthetic datasets for training and evaluating computer vision algorithms that are to be deployed in the real world. While the visual fidelity of modern game engines has been significantly improved with technologies such as ray-tracing, a notable sim2real appearance gap between the synthetic and the real-world images still remains, which limits the utilization of synthetic datasets in real-world applications. In this letter, we investigate the ability of a state-of-the-art image generation and editing diffusion model (FLUX.2-4B Klein) to enhance the photorealism of synthetic datasets and compare its performance against a traditional image-to-image translation model (REGEN). Furthermore, we propose a hybrid approach that combines the strong geometry and material transformations of diffusion-based methods with the distribution-matching capabilities of image-to-image translation techniques. Through experiments, it is demonstrated that REGEN outperforms FLUX.2-4B Klein and that by combining both FLUX.2-4B Klein and REGEN models, better visual realism can be achieved compared to using each model individually, while maintaining semantic consistency. The code is available at: \url{https://github.com/stefanos50/Hybrid-Sim2Real}
\end{abstract}


\section{Introduction} \label{sec:intro}

Video game engines have emerged as a promising approach for generating large-scale visual synthetic datasets \cite{10091796} for training and evaluating Computer Vision (CV) algorithms \cite{yolov12, mask2former}. In detail, their ability to automatically produce accurate annotations in fully controllable and customizable environments has made them an attractive alternative in scenarios where the generation of real-world datasets is time-consuming, costly, or unsafe. Despite the significant progress in technologies that are incorporated in modern game engines (e.g., Unreal Engine 5), such as Lumen and Nanite, a noticeable visual gap between the synthetic and the real-world images, often referred to as the simulation to reality (sim2real) appearance gap \cite{carla2real,regen}, persists. This sim2real appearance gap limits the real-world applicability of CV algorithms trained solely on synthetic datasets, as they fail to achieve an adequate level of generalization on the real-world visual characteristics and complexities.

To reduce the sim2real appearance gap, most approaches focus on either Image-to-Image (Im2Im) translation \cite{regen, epe, carla2real} or diffusion-based \cite{sim2realdiffusion, wang2025zeroshotsyntheticvideorealism} methods. In detail, Im2Im translation methods are effective at enhancing the photorealism of synthetic datasets by translating their visual characteristics towards the ones of a target real-world dataset \cite{cityscapes,kitti} while achieving real-time inference, as well as temporal and semantic consistency \cite{regen}. On the other hand, diffusion-based methods enable zero-shot photorealism enhancement \cite{wang2025zeroshotsyntheticvideorealism} guided by textual prompts and can achieve high levels of visual realism by performing strong geometry and material changes. However, both approaches can be subject to various limitations in reducing the sim2real appearance gap. Im2Im translation methods, while effective at transferring the distribution and characteristics of the target real-world dataset \cite{carla2real}, tend to perform fewer geometry and material updates in order to preserve semantic consistency \cite{epe}, which limits the achievable realism, especially when low-quality synthetic objects are depicted (e.g., with a low amount of polygons or triangles). Diffusion-based methods are prone to frequent hallucination, even when multiple control signals (e.g., depth and edge) are employed \cite{wang2025zeroshotsyntheticvideorealism, regen}, resulting in the photorealism-enhanced images deviating from the ground-truth annotations (e.g., annotated bounding boxes for object detection). In addition, compared to Im2Im translation, which is explicitly trained to match the distribution of a synthetic dataset with that of a target real-world dataset, diffusion-based methods struggle to accurately reflect the diversity and complexities of the real-world data distributions \cite{alimisis2025advancesdiffusionmodelsimage}. As a result, the contribution to the improvement of the real-world generalization performance of CV algorithms trained on synthetic datasets that are photorealism-enhanced by diffusion-based methods (reduction of the sim2real appearance gap) may be limited. 

In this letter, considering the aforementioned limitations of both Im2Im translation and diffusion-based methods, we examine the performance of a recent (January 2026) State-of-The-Art (SoTA) image generation diffusion model with strong editing capabilities, namely, FLUX.2-4B Klein \cite{flux4b_klein}, for photorealism enhancement of synthetic datasets, and compare it with the most recent (February 2026) SoTA Im2Im translation photorealism enhancement model, REGEN \cite{regen}. In addition, we propose a hybrid approach that combines diffusion (i.e., FLUX.2-4B Klein) and Im2Im translation (i.e., REGEN) to enhance the photorealism of synthetic datasets generated from game engines. Through experiments on synthetic datasets extracted from the Unity and the Rockstar Advanced Game Engine (RAGE) game engines using a metric that has been proven to align with human judgment, namely, CLIP Maximum Mean Discrepancy (CMMD) \cite{cmmd}, it is illustrated that the effective translation towards the distribution and characteristics of real-world data by REGEN is more important than the strong image editing capabilities of FLUX.2-4B Klein (e.g., geometry changes) and that with their combination more photorealistic images of the synthetic datasets can be produced compared to applying each model (FLUX.2-4B Klein or REGEN) individually. In addition, it is shown that these photorealism-enhanced images remain faithful to the ground-truth annotations of the synthetic datasets using pretrained semantic segmentation (i.e., Mask2Former \cite{mask2former}) and object detection (i.e., YOLO26 \cite{yolov12}) models.

\begin{figure}[htbp]
    \centering
    \includegraphics[width=0.5\textwidth]{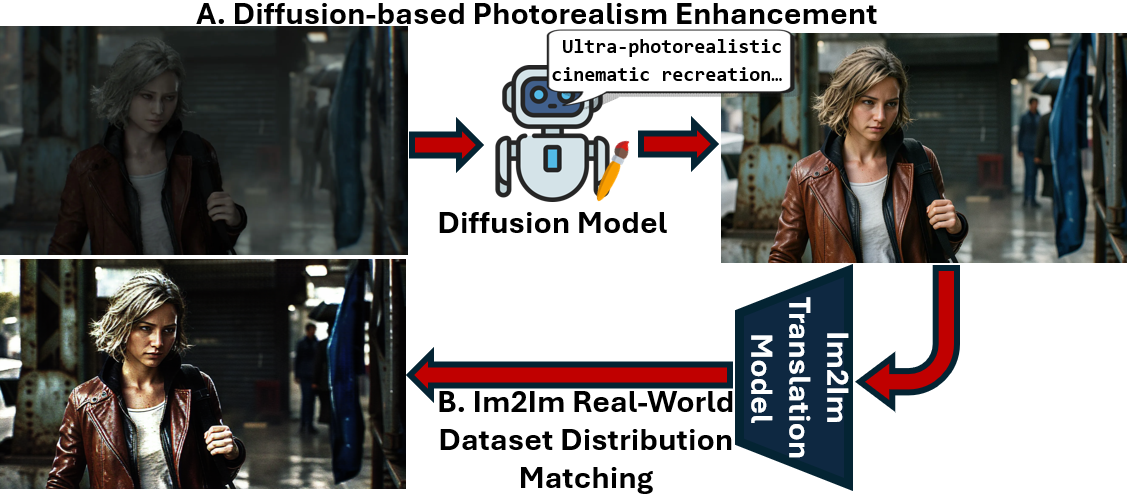}
    \caption{Overview of the proposed hybrid photorealism-enhancement approach, which is split into two phases: a) the diffusion-based photorealism enhancement and b) the im2im real-world dataset distribution matching phase. The example input image is from the Resident Evil Requiem video game.}
    \label{fig:flowchart}
\end{figure}

\section{Photorealism Enhancer}

In this section, the proposed hybrid photorealism-enhancement approach is detailed, as illustrated in Fig. \ref{fig:flowchart}, which includes two phases: (i) the diffusion-based photorealism enhancement, and (ii) the Im2Im real-world dataset distribution matching phase.

\subsection{Diffusion-based Photorealism Enhancement}

In the diffusion-based photorealism enhancement phase, a synthetic image generated from a game engine is processed by a diffusion-based method to produce its photorealism-enhanced counterpart. Specifically, we employ FLUX.2-4B Klein \cite{flux4b_klein}, which is one of the most lightweight image generation and editing diffusion models that can be inferred in consumer-grade hardware (e.g., NVIDIA RTX 3090) as it has a requirement of roughly 13GB of VRAM. Moreover, FLUX.2-4B Klein has no requirement for additional control signals (e.g., semantic segmentation maps) \cite{wang2025zeroshotsyntheticvideorealism} that often limit the applicability of such models on pre-existing synthetic datasets that were not exported with this information (FLUX.2-4B Klein requires only an RGB image as input). Finally, FLUX.2-4B Klein was selected as it has strong image editing capabilities and therefore can enhance the photorealism of lighting, geometry, and materials while preserving the initial synthetic image structure and layout. This is particularly an important factor for CV algorithms, as photorealism enhancement must preserve alignment with the ground truth annotations. Any hallucinated or distorted objects introduce mismatches (e.g., between the images and the semantic segmentation maps of a dataset) that will subsequently negatively affect the performance of the CV algorithms during training.

\subsection{Im2Im Real-World Dataset Distribution Matching}

In the Im2Im real-world dataset distribution matching phase, the photorealism-enhanced image produced by the diffusion-based method (i.e., FLUX.2-4B Klein) is fed to an Im2Im translation method that is trained to translate an input image towards the distribution and characteristics of a real-world dataset that was employed as the target of realism during training. Therefore, the trained Im2Im translation model adapts the photorealism-enhanced image produced by the diffusion-based method by adding the complexities and characteristics of a specific real-world dataset. As a result, the synthetic data becomes closer to real-world data distribution, and thus, the sim2real appearance gap is further reduced. In detail, to perform this, we select a SoTA Im2Im translation model for photorealism enhancement, REGEN \cite{regen}, which learns to regenerate the output of a robust Im2Im translation model \cite{epe}, removing the requirement of additional inputs (e.g., depth) and improving the inference time. REGEN requires as input solely an RGB synthetic image, and therefore can be applied to any pre-existing synthetic dataset. REGEN is provided with two models that were trained to translate the CARLA simulator towards the characteristics of the KITTI \cite{kitti} and Cityscapes (CS) \cite{cityscapes} real-world datasets. Finally, REGEN was proven to maintain semantic and temporal consistency.

\section{Experiments and Discussion}

\subsection{Synthetic Datasets and Metrics}

\paragraph{Synthetic Datasets} Two datasets that were extracted from two different game engines are employed for the experiments. Virtual KITTI 2 (VKITTI2) \cite{cabon2020virtualkitti2} is a dataset that clones five scenes of the real-world KITTI dataset, including a total of $2,126$ images. VKITTI2 was generated from the Unity game engine with a dash cam perspective and includes annotations, such as semantic segmentation maps (15 object categories) and camera intrinsics. In addition, a Roboflow dataset\footnote{\url{https://universe.roboflow.com/ilhamfazri3rd-gmail-com/gta-v-vehicle-dataset}} generated with an Aerial Unmanned Vehicle (UAV) perspective (DeepGTA tool \cite{deepgta}) from the video game Grand Theft Auto V (GTA-V) that is based on the RAGE game engine is utilized in the experiments. Specifically, the dataset includes $456$ images accompanied by bounding box annotations for object detection (5 object categories).

\paragraph{Metrics} To evaluate visual realism, the CMMD \cite{cmmd} metric is employed, which evaluates the similarity between a reference (i.e., real-world) and a generated (i.e., synthetic or photorealism-enhanced) dataset. In detail, CMMD was selected since it has been proven to align with human perception and judgment through user studies. In addition, to assess whether photorealism-enhanced images remain faithful to the structure and layout of the initial synthetic images of synthetic datasets, Intersection over Union (IoU) is used for evaluating the predictions of a semantic segmentation, and mean Average Precision at IoU threshold 0.50 (mAP@50) for an object detection model, compared to the ground-truth annotations.

\begin{figure*}[htbp]
    \centering
    \includegraphics[width=0.95\textwidth]{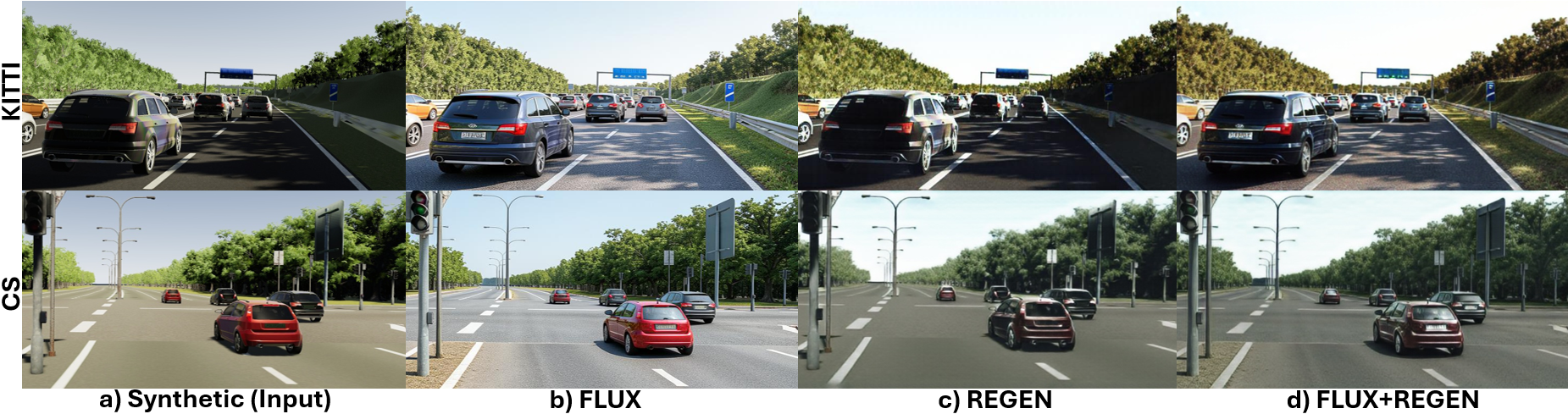}
    \caption{Visual examples of the photorealism-enhanced image produced by b) FLUX, c) REGEN, d) FLUX+REGEN, given a) an input from the VKITTI2 dataset for two real-world dataset variations, KITTI and CS.}
    \label{fig:visual_example}
\end{figure*}

\subsection{Experimental Setup}

To conduct the experiments, FLUX.2-4B and REGEN were employed. Both models are pretrained and have never seen during training the selected datasets (i.e., VKITTI2 and GTA-V). In more detail, FLUX.2-4B is a foundation diffusion model that is designed to perform zero-shot image generation, and REGEN is trained on synthetic images from the CARLA simulator \cite{carla2real}, which were generated from the Unreal Engine 4 game engine. Along those lines, for VKITTI2 ($2,126$ images), first FLUX.2-4B Klein was prompted (see Fig. \ref{fig:prompt}) to enhance the photorealism of the images without altering the layout of the scene (from now on defined as FLUX). Next, REGEN, trained to translate CARLA towards the KITTI and CS real-world characteristics, was applied to the synthetic images of VIKITTI2 (defined as REGEN) and subsequently to the photorealism-enhanced images (output) of FLUX (defined as FLUX+REGEN). No resizing was applied to the images. The same process was followed for the GTA-V dataset ($456$ images), resulting in two variations (KITTI and CS) for REGEN and two variations for FLUX+KITTI. To evaluate the visual realism and the reduction of the sim2real appearance gap, first, the synthetic images (defined as Synthetic) of the datasets (VKITTI2 and GTA-V) are evaluated against the real-world KITTI and CS datasets using CMMD. For the real-world KITTI, the exact $2,126$ clone images from VKITTI2 are selected, and for CS, the entire $5,000$ images of the dataset. Then, the KITTI variations of REGEN and FLUX+REGEN are evaluated against KITTI, and the CS variations of REGEN and FLUX+REGEN against the CS dataset.

To evaluate semantic preservation, pretrained CV models are applied on the synthetic images (Synthetic) and the photorealism-enhanced ones of the synthetic datasets (VIKITTI2 and GTA-V) produced by the best-performing photorealism enhancement approach (i.e., FLUX+REGEN). For VIKITTI2, where semantic segmentation maps are available, the official pretrained Mask2Former model trained on the CS dataset was employed. To enable compatibility with the CS object categories, the tree and vegetation categories were merged, as well as the truck with van, and misc with unlabeled, with the latter not considered for evaluation since it doesn't exist in the pretrained model (11 total object categories). Particularly, the Mask2Former \cite{mask2former} was first applied and evaluated (using mIoU) on the synthetic images (Synthetic) and then on the two variations (KITTI and CS) of FLUX+REGEN. Maintaining a similar mIoU between the synthetic images and the photorealism-enhanced ones illustrates that the photorealism-enhanced images are semantically consistent. For GTA-V (5 object categories), a pretrained YOLO26m \cite{yolov12} object detector is applied on the synthetic images and on the two variations (KITTI and CS) of FLUX+REGEN, and mAP@50 is calculated. Again, similar mAP@50 between the synthetic and the photorealism enhanced images indicates semantic consistency.

\subsection{Results and Discussion}

In this section, the synthetic images of VIKITTI2 and GTA-V, the photorealism-enhanced images produced by FLUX, as well as the two variations towards the characteristics of the CS and KITTI real-world datasets of REGEN and FLUX+REGEN, are evaluated in terms of visual realism against the respective real-world datasets using CMMD (lower is better). In addition, a Mask2Former segmentation model is applied on the synthetic and the variations (CS and KITTI) of FLUX+REGEN for VIKITTI2 and a YOLO26m object detection model for GTA-V in order to evaluate semantic consistency with mIoU and mAP@50 metrics, respectively (higher is better). Along those lines, Table \ref{tab:vkittti_gta_results} presents the visual realism comparison using the CMMD metric. It is evident that, between FLUX and REGEN in most of the cases, REGEN leads to a more significant CMMD reduction (lower CMMD indicates higher similarity) compared to FLUX. This highlights that the distribution and characteristics matching with the target real-world dataset is more important compared to the strong geometry and material changes introduced by FLUX. However, when combining (FLUX+REGEN), the increased geometry and material updates of FLUX with the distribution matching of REGEN, it is illustrated that across all evaluation cases, the visual realism is increased (i.e., CMMD is further reduced). Particularly, for the VIKITTI2 dataset, applying FLUX to increase the photorealism in terms of geometry and materials of the low-quality objects depicted in the images of the dataset and subsequently transforming it towards the distribution of the KITTI dataset using REGEN leads to a notable low CMMD value of $1.781$ (reduction of the sim2real appearance gap), which indicates that the clone synthetic images are significantly close to the real-world ones in terms of visual realism. This is visually depicted in Fig. \ref{fig:visual_example}, where it is evident that FLUX performs significant geometry and material changes, REGEN transforms towards the distribution and characteristics of the real-world datasets (e.g., dark color distribution for CS), and their combination results in photorealistic images that include both aspects (improved geometry and materials as well as the real-world distribution and complexities). Finally, regarding the semantic consistency of the photorealism-enhanced images, as shown in Table \ref{tab:miou_map_results}, the accuracy (mIoU) of Mask2Former on the VKITTI2 dataset not only is matched between the photorealism-enhanced and the synthetic images, but it is instead increased on the photorealism ones due to the better feature alignment. As expected, since the model is trained on the CS real-world dataset, the highest mIoU is achieved on the FLUX+REGEN towards the CS characteristics. For the YOLO26m model on the GTA-V dataset, the mAP@50 remains similar between the synthetic and the KITTI and CS variations of FLUX+REGEN, which again highlights semantic consistency.

\begin{figure}[t]
\centering
\begin{tcolorbox}[colback=gray!5,colframe=black,boxrule=0.5pt,
                  arc=2pt,left=4pt,right=4pt,top=4pt,bottom=4pt]
\scriptsize
Ultra-photorealistic cinematic recreation of the input image. Preserve the exact composition, camera angle, framing,
perspective, scale, and spatial placement of every object — no additions, removals, or repositioning. Replace any
CGI/game aesthetics with real-world materials featuring physically accurate textures (micro-details, imperfections,
wear, fingerprints, surface variation). Use physically based rendering with correct light transport, global illumination,
realistic reflections, refractions, and contact shadows. Natural color grading, high dynamic range, true-to-life skin
tones (if applicable), and accurate material response (metal, glass, fabric, skin, etc.). Shot as if captured with a
high-end cinema camera: shallow depth of field where appropriate, natural lens characteristics, subtle film grain,
and cinematic lighting. No stylization, no painterly effects, no artificial smoothing, no geometry changes — pure
real-world realism.
\end{tcolorbox}
\caption{Prompt used for photorealism-enhancement with FLUX.2-4B Klein.}
\label{fig:prompt}
\end{figure}

\subsection{Limitations}

The primary limitation is that diffusion-based methods, even the ones that are designed for videos \cite{wang2025zeroshotsyntheticvideorealism}, are still subject to temporal inconsistencies, which limit their applicability to sequential visual data (e.g., videos). As a result, the approach is applicable for synthetic datasets for frame-level tasks such as image classification, object detection, semantic segmentation, and depth estimation. In addition, since the approach relies on a diffusion-based method, it cannot be applied in real-time \cite{regen} (e.g., simulations). However, with the release of NVIDIA's Deep Learning Super Sampling 5.0 (DLSS 5.0), these limitations can be potentially addressed (i.e., using DLSS 5.0 combined with REGEN).

\section{Conclusions}

In this letter, the capability of the FLUX.2-4B Klein image generation and editing diffusion model to perform photorealism-enhancement of synthetic datasets was investigated and compared against a traditional Im2Im translation model, REGEN. In addition, a new approach was proposed that utilizes the strong geometry and material changes of diffusion-based methods with the distribution and characteristic matching provided by the Im2Im translation methods. Through experiments, it was demonstrated that matching the real-world dataset distribution is more important in closing the sim2real appearance gap, resulting in REGEN outperforming FLUX.2-4B Klein model, while the introduced hybrid approach that combines both models was proven to lead to better visual realism and to produce semantically consistent photorealism-enhanced images.

\begin{table}[h!]
\centering
\caption{Visual realism comparison of Synthetic, FLUX, REGEN, and FLUX+REGEN on VKITTI2 and GTA-V against the KITTI and CS real-world datasets using CMMD (lower is better).}
\resizebox{0.8\columnwidth}{!}{
\begin{tabular}{lcc|cc}
\hline
Model & \multicolumn{2}{c|}{VKITTI2} & \multicolumn{2}{c}{GTA-V} \\
& KITTI & CS & KITTI & CS \\
\hline
Synthetic      & 3.734 & 4.805 & 6.321 & 6.333 \\
\hdashline
FLUX           & 2.488 & 4.561 & 5.674 & 5.332 \\
REGEN          & 2.726 & 3.923 & 4.730 & 4.861 \\
FLUX+REGEN   & 1.781 & 3.751 & 3.956 & 4.326 \\
\hline
\end{tabular}
}
\label{tab:vkittti_gta_results}
\end{table}

\begin{table}[h!]
\centering
\caption{Semantic consistency comparison between the synthetic and the FLUX+REGEN CS and KITTI variations on the VIKITTI2 and GTA-V using mIoU and mAP@50 (higher is better), respectively.}
\resizebox{1\columnwidth}{!}{
\begin{tabular}{lcc}
\hline
Model & VKITTI2 (mIoU) & GTA-V (mAP@50) \\
\hline
Synthetic & 52.18\% & 48.20\% \\
\hdashline
FLUX+REGEN (KITTI)     & 53.41\% & 49.10\% \\
FLUX+REGEN (CS)        & 55.94\% & 47.70\% \\
\hline
\end{tabular}
}
\label{tab:miou_map_results}
\end{table}

\bibliographystyle{IEEEtran}
\bibliography{bibliography}

\vfill

\end{document}